\documentclass[sigconf, nonacm]{acmart}

\usepackage{multirow}
\usepackage{graphicx}
\usepackage{subcaption}
\usepackage{hyphenat}
\begin{document}

\title{FaceInsight: A Multimodal Large Language Model for Face Perception}


\author{Jingzhi Li$^{1,2}$, Changjiang Luo$^{1,2}$, Ruoyu Chen$^{1,2}$, Hua Zhang$^{1,2}$, Wenqi Ren$^{3}$, Jianhou Gan$^{4}$, \\ Xiaochun Cao$^{3}$}
\affiliation{$^{1}$ Institute of Information Engineering, Chinese Academy of Sciences \country{}}
\affiliation{$^{2}$ School of Cyber Security, University of Chinese Academy of Sciences \country{}}
\affiliation{$^{3}$ School of Cyber Science and Technology, Shenzhen Campus of Sun Yat-sen University \country{}}
\affiliation{$^{4}$ Key Laboratory of Education Informatization for Nationalities (Yunnan Normal University), Ministry of Education \country{}}

\begin{abstract}

Recent advances in multimodal large language models (MLLMs) have demonstrated strong capabilities in understanding general visual content. However, these general-domain MLLMs perform poorly in face perception tasks, often producing inaccurate or misleading responses to face-specific queries. To address this gap, we propose FaceInsight, the versatile face perception MLLM that provides fine-grained facial information. Our approach introduces visual-textual alignment of facial knowledge to model both uncertain dependencies and deterministic relationships among facial information, mitigating the limitations of language-driven reasoning. Additionally, we incorporate face segmentation maps as an auxiliary perceptual modality, enriching the visual input with localized structural cues to enhance semantic understanding. Comprehensive experiments and analyses across three face perception tasks demonstrate that FaceInsight consistently outperforms nine compared MLLMs under both training-free and fine-tuned settings.

\end{abstract}

%





\begin{teaserfigure}
	\centering
	\begin{subfigure}[b]{0.7\textwidth}
		\includegraphics[width=\textwidth]{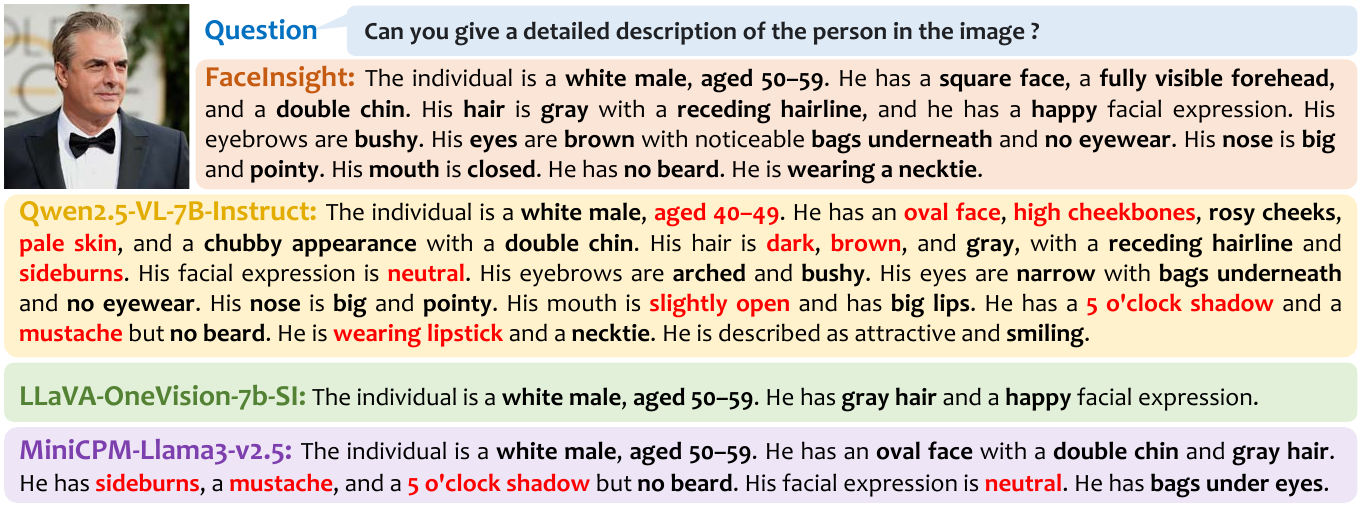}
	\end{subfigure}
	\begin{subfigure}[b]{0.29\textwidth}
		\includegraphics[width=\textwidth]{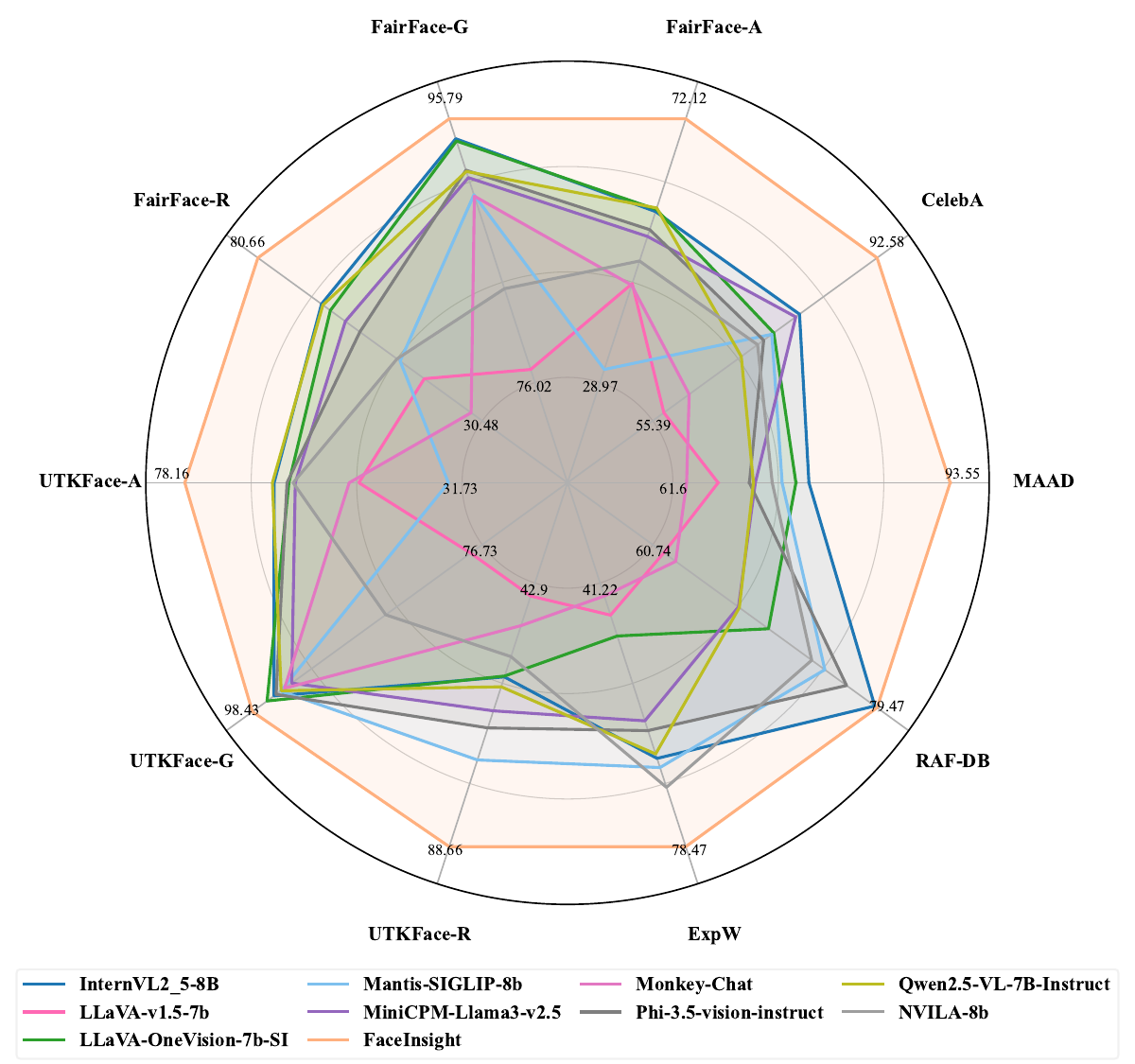}
	\end{subfigure}
    \vspace{-10pt}
	\caption{Performance of Multimodal Large Models (MLLMs) on Face Perception Tasks. The left panel shows qualitative results, with bold indicating correctly identified features and red marking misidentified ones. The right panel presents the accuracy of FaceInsight and nine MLLMs across face perception tasks. The letters following each dataset name denote specific tasks: A for age estimation, G for gender estimation, and R for race estimation. Best viewed zoomed-in.}
	\label{fig_intro}
\end{teaserfigure}


\maketitle

\section{Introduction}


Faces provide rich, multifaceted information about individuals, enabling the inference of key attributes such as identity, expression, age, gender, ethnicity, and attractiveness. The ability to accurately perceive and interpret facial information is fundamental to a wide range of real-world, human-centric applications. As multimodal large language models (MLLMs) are increasingly integrated into intelligent systems such as virtual reality, autonomous driving, human-computer interaction, authentication, and healthcare, it is crucial that these models possess robust face perception capabilities. In these contexts, facial information is not only frequently encountered but also critical for generating appropriate and context-aware responses.

Despite rapid advancements in MLLMs on vision-language tasks such as captioning, visual reasoning, and image generation, their capabilities in face perception remain limited.
The general-purpose MLLMs often struggle to answer even basic facial queries, such as recognizing expressions or localizing facial regions, mainly because they are trained on broad-domain image-text pairs and rely excessively on language-driven reasoning.
These limitations significantly hinder their effectiveness in applications requiring nuanced facial interpretation. A recent study, FaceXBench \cite{narayan2025facexbench}, systematically evaluates multiple MLLMs, highlighting their deficiencies in facial understanding. Furthermore, these models frequently exhibit hallucinations and logical inconsistencies when processing facial content, as illustrated in Fig. \ref{fig_intro}. The understanding of MLLMs is often driven more by linguistic patterns than by visual cues.

Face perception requires moving beyond general visual understanding to capture the fine-grained, multimodal nature of facial cues—including subtle expressions, localized features, and dynamic variations.
Recent research has begun to explore this direction. Face-MLLM \cite{sun2024face} proposed a three-to enhance the facial perception capabilities of MLLMs. FaVChat \cite{zhao2025favchat} focused on fine-grained comprehension of facial video content. EMO-LLaMA \cite{xing2024emo} introduced an instruction-tuning dataset aimed at enhancing expression understanding.
However, current efforts concentrate on constructing supervised datasets, with limited attention given to aligning model knowledge specifically for facial perception tasks \cite{narayan2025facexbench}. We believe that visual-text correspondence alone is insufficient in this context, as a single facial region often encodes multiple overlapping attributes—such as a smile, lipstick, or a slightly open mouth—necessitating more sophisticated modeling approaches.

In this paper, we propose FaceInsight, an MLLM-based framework explicitly designed to enhance face perception. We conduct a thorough investigation, highlighting the integration of domain-specific facial knowledge into the MLLM framework, and propose targeted solutions. Specifically, we propose adding an additional perceptual modality—face segmentation maps—into the framework. This provides additional localized visual information, embedding spatial-aware facial visual knowledge into the MLLM. Subsequently, we design a correlation constraint module to explore the dependencies among facial features, guided by the constructed correlation maps. And a logical constraint module is introduced to thoroughly investigate the explicit logical relationships among facial features.
This approach not only addresses the limitations of visual encoders, which often fail to comprehensively extract the visual information necessary for face perception, but also reduces the reliance on language-based memory in large models.

To evaluate the effectiveness of our method, we perform comprehensive experiments across three face perception tasks, utilizing six datasets and comparing nine MLLMs. The results consistently demonstrate that our method outperforms state-of-the-art approaches across all datasets, showing significant improvements and confirming its superiority.

In summary, our main contributions are as follows:
\begin{itemize}
	\item We introduce FaceInsight, a novel face perception framework that exploits the advanced visual representation learning capabilities of multimodal LLMs. FaceInsight effectively integrates visual and textual knowledge to enhance the fine-grained understanding of facial images in MLLMs while reducing their reliance on language-based memory.
	\item We design a correlation constraint module and a logical constraint module to comprehensively capture facial semantic knowledge, encompassing both uncertain dependencies and certain logical relationships, thereby ensuring that the model's face perception aligns with intrinsic facial correlations.
	\item Experimental results show that our method consistently achieves the best performance on all benchmark datasets, highlighting its remarkable effectiveness. In-depth analysis further demonstrates the superiority of our method.
\end{itemize}

\section{Related Work}

\subsection{Visual Understanding with MLLMs}

Recent advances in multimodal large language models (MLLMs) have driven progress in visual understanding \cite{awadalla2023openflamingo, liu2023visual, liu2024llava,li2025visual}.
The typical architecture includes a pre-trained visual backbone \cite{radford2021learning, jain2024vcoder} for encoding visual data, a cross-modal projector—such as a linear projection layer or multilayer perceptron (MLP)—to align visual and textual features, and a large language model \cite{chiang2023vicuna, qi2024cogcom} that functions as the core intelligence engine, interpreting user instructions and generating responses.

LLaVA \cite{liu2023visual} pioneered visual-language instruction tuning by converting image-text data into dialog format and fine-tuning CLIP \cite{radford2021learning} and LLaMA \cite{touvron2023llama} on this dataset. MiniGPT-4 \cite{zhu2023minigpt} similarly leveraged BLIP2’s visual encoder \cite{li2023blip} with a linear projection to align with Vicuna’s \cite{chiang2023vicuna} feature space. InstructBLIP’s \cite{instructblip} Q-Former and projection layer further strengthened LLMs’ instruction-following capabilities.
Other notable frameworks include mPLUG-Owl \cite{ye2023mplug}, which introduced a vision abstractor, and LLaVA-1.5 \cite{liu2024improved}, which refined MLLM performance with an MLP projector. LION \cite{chen2024lion} integrated dual-level visual knowledge to handle task conflicts, while MiniGPT-v2 \cite{chen2023minigpt} used task identifiers to improve instruction distinction and learning efficiency.
Expanding to video and multi-image comprehension, VILA \cite{lin2024vila} employed large-scale interleaved pretraining, and MoE-LLaVA \cite{lin2024moe} enhanced efficiency with a mixture-of-experts design, selectively activating experts for optimized visual understanding.
InternVL2.5-8B \cite{chen2024expanding} improves training strategies and multilingual reasoning, achieving good performance on the MMMU benchmark. MiniCPM-Llama3-v2.5 \cite{yao2024minicpm} is optimized for mobile deployment, featuring high-resolution vision, low hallucination rates, and multilingual support.  Mantis-SIGLIP-8B \cite{jiang2024mantis} is presumed to enhance image-text alignment using the SIGLIP framework. Monkey-Chat \cite{li2024monkey}, Phi-3.5-Vision-Instruct \cite{abdin2024phi}, and Qwen2.5-VL-7B-Instruct \cite{bai2025qwen2} are designed for multimodal instruction-following tasks, though detailed information on their architectures and capabilities remains limited.

While MLLMs excel in vision-language tasks such as image captioning, visual reasoning, and image generation, their face perception capabilities remain limited. We identify two key challenges: (1) the complexity of facial information within a constrained visual area. Multiple features such as smiling, lipstick, or mouth slightly open coexist in a single region, making simple visual-text alignment insufficient for accurate interpretation. (2) MLLMs’ heavy reliance on language-based memory, which can lead to hallucinations or reasoning errors, as their responses are often shaped more by linguistic patterns than visual cues (see Fig. \ref{fig_intro}).


\vspace{-20pt}

\subsection{Face Perception Model}

Face perception refers to the automated processes by which algorithms detect, analyze, and interpret human facial features. Significant advancements in this field have led to state-of-the-art methods across various tasks, including facial attribute recognition \cite{hassanat2022deepveil, yan2023spl, wu2023logical, wu2023logicnet}, expression and emotion recognition \cite{canal2022survey, cheng2023semi,li2024facial}, and age, gender, and race estimation \cite{cao2020rank, li2021learning, kuprashevich2023mivolo}. Concurrently, some studies have developed unified models capable of performing multiple face perception tasks. Early works, such as HyperFace \cite{ranjan2017hyperface} and AllinOne \cite{ranjan2017all}, used CNN backbones with empirically selected feature layers, enabling them to detect landmarks, estimate head pose and gender, and in AllinOne, conduct face recognition and age estimation. Recently, unified models based on Transformer architectures—such as SwinFace \cite{qin2023swinface}, FaceXformer \cite{narayan2024facexformer}, Qface \cite{sun2024task}, and Faceptor \cite{qin2025faceptor}—have emerged as powerful alternatives.

In addition, visual-language models have shown promise in face perception tasks, where natural language supervision enhances visual representation learning. FaRL \cite{zheng2022general} employs image-text contrastive learning and masked image modeling to create generalized facial representations, excelling in facial attribute recognition tasks. Bulat et al. \cite{bulat2022pre} explored unsupervised pre-training for facial representation learning, yielding benefits across multiple facial analysis applications. Label2Label \cite{li2022label2label} improves facial attribute recognition by employing an image-conditioned masked language model, which predicts randomly masked “words” using contextual image information. Recently, DFER-CLIP \cite{zhao2023prompting}, based on CLIP, was designed for dynamic facial expression recognition in diverse settings, while EmoCLIP \cite{foteinopoulou2024emoclip} enhances latent representations for zero-shot facial expression recognition through sample-level text descriptions.

Recent research has begun to explore the potential of multimodal large models in the domain of face perception. Face-MLLM \cite{sun2024face} proposed a three-stage training pipeline to improve MLLMs' facial perception. FaVChat \cite{zhao2025favchat} focused on fine-grained comprehension of facial video content. EMO-LLaMA \cite{xing2024emo} introduced a dataset aimed at enhancing expression understanding. FaceXBench \cite{narayan2025facexbench}, systematically evaluates multiple MLLMs, highlighting their deficiencies in facial understanding.
Despite these efforts, existing methods remain constrained by several challenges, including inadequate attention to facial cues and logical consistency, as well as limited generalizability beyond specific benchmarks. We argue that relying solely on visual-text correspondence is insufficient for face perception tasks, highlighting the need for more sophisticated and context-aware modeling approaches.


\begin{figure*}[!t]
    \centering
    \includegraphics[width=1\textwidth]{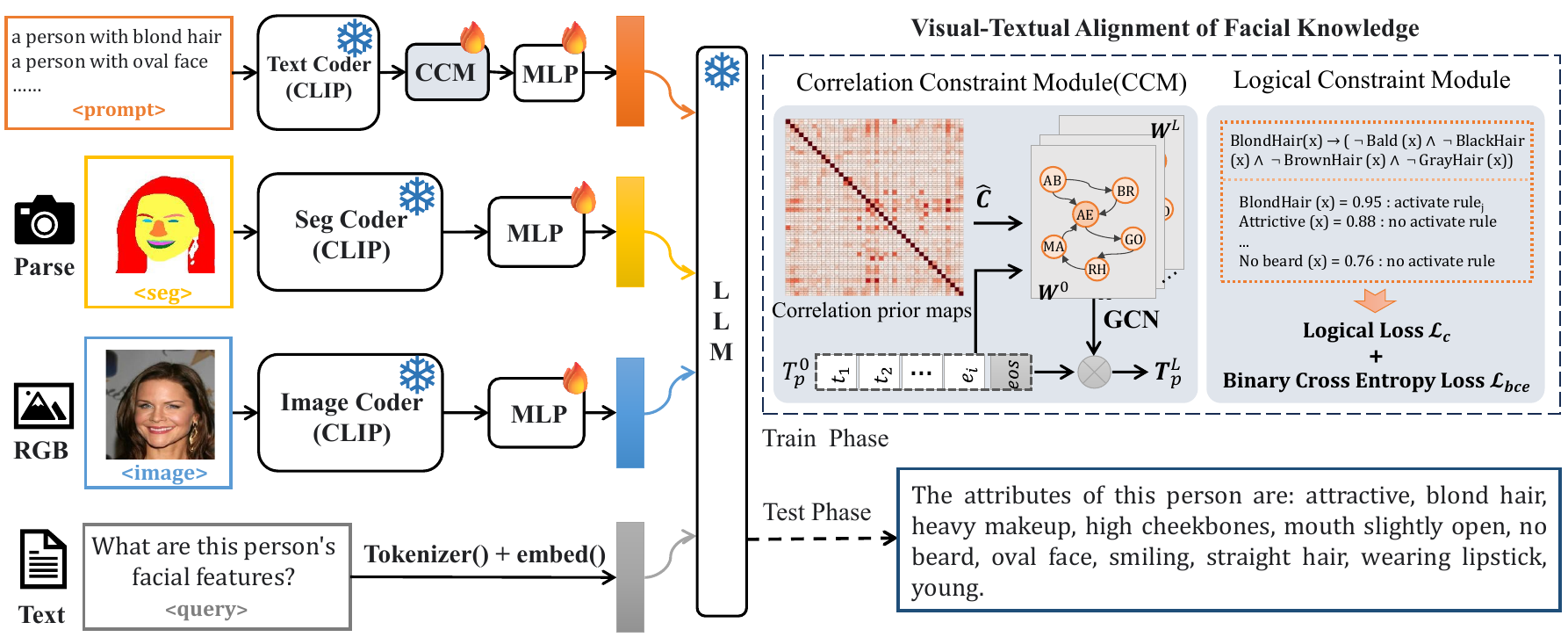}
    \vspace{-18pt}
    \caption{ An overview of FaceInsight. To fully integrate fine-grained knowledge of facial information, we design a correlation constraint module and a logical constraint module to model uncertain dependencies and certain logical relationships, respectively. Additionally, we incorporate the face segmentation modality to provide region-level visual information, embedding spatial-aware facial visual knowledge into the framework.}
    \label{fig_framework}
\end{figure*}

\section{Methodology}

Visual conversations for face perception aim to generate text responses linked to specific visual content, providing fine-grained analysis of facial features. This includes global characteristics such as gender, age, and expression, along with local features like hair color and beard type. Although MLLMs demonstrate capabilities across various visual-language tasks, including visual captioning, image generation, and visual reasoning, they often perform poorly in face perception tasks. As illustrated in Fig. \ref{fig_intro}, when prompted to provide detailed descriptions of faces in the visual input, general-domain MLLMs output incorrect information or introduces hallucinations (a logical conflict between 'no beard' and 'goatee').

Through our investigation of these MLLMs, we identify two primary reasons for this phenomenon. First, the complexity and richness of facial information within a constrained visual area contribute to challenges. These MLLMs extract visual data through pre-trained encoders focused primarily on texture, structure, and prominent features, as their training based on image captions. This limited focus could result in insufficient facial detail, causing MLLMs to generate inaccurate or hallucinatory information. Second, MLLMs' inherent reliance on language-based memory often leads to inaccurate outputs. As shown in Fig. \ref{fig_intro}, when different initial descriptions of facial features are provided, the model's responses vary significantly.

Inspired by these findings, we develop FaceInsight to enhance MLLMs' fine-grained understanding of facial images by integrating visual and textual knowledge. First, we add face segmentation maps to provide additional localized visual information, embedding spatial-aware facial visual knowledge into the MLLM. Then, we design a correlation constraint module and a logical constraint module to model uncertain dependencies and certain logical relationships among facial features, respectively, embedding domain-specific expertise into the framework. This design not only compensates for the shortcoming of visual encoders that are insufficient to comprehensively extract the visual information required for face perception, but also mitigates the reliance on language-based memory in large models.

\subsection{FaceInsight Architecture}

The multimodal LLMs take multimodal information as input, including an image $I$ and text $T$ of random length.The image $I$ is encoded using a image encoder $E_i$, allowing it to share the same embedding space as the text embedding. The text can be further divided into two parts: the first part is the user query $T_q$, which relates to the user's requirements, while the second part is the pre-designed task prompt $T_p$, which is either hand-crafted or generated automatically. Although the prompt is not provided by the user, an appropriate prompt can guide the MLLMs to generate a higher-quality response.
Formally, the multimodal LLMs $M_{llm}$ generate the answer $T_a$ by
\begin{equation}
	T_a = M_{llm}(E_i(I), T_q, T_p).
\end{equation}

In this paper, we add face segmentation maps $I_s$ to incorporate shape perception as a modality, providing additional image-specific information for face perception. We also add the correlation prior maps $C$ to refine the text prompt $T_p$, embedding the dependencies among facial characteristics. Our model can be formulated as:
\begin{equation}
	T_a = M_{llm}(E_i(I), E_s(I_s), T_q, A(T_p, C)),
\end{equation}
where $E_s(\cdot)$ and $A(\cdot)$ indicate the segmentation map encoder and the correlation constraint module, respectively.

As shown in Fig. \ref{fig_framework}, FaceInsight follows the architecture of LLaVA-v1.5 \cite{liu2024improved} and consists of three core components: multimodal encoder, MLP adapter (comprising two linear layers), and LLM.
For the input image, we adopt the Vision Transformer (ViT) from CLIP \cite{radford2021learning}, pre-trained on the 80M face image dataset FLIP, as the image encoder to ensure greater consistency with face-related tasks.
Then, we utilize face segmentation maps from Pyfacer \cite{deng2020retinaface} and employ a pretrained ViT as the segmentation encoder. The input image and the segmentation map are converted into a fixed number of patch embeddings.
For text prompt inputs, we first use the pretrained text encoder from CLIP to extract text features. To capture the fine-grained knowledge of facial information, we design a correlation constraint module, which uses prior knowledge of facial features to gradually optimize the textual representation. Specifically, we input the initial text features together with the semantic knowledge maps into a graph convolutional neural network to obtain the final textual embedding.
To process multimodal prompts beyond queries, we adopt two-layer MLP to project the encoded features into the language space.
Both embeddings are then projected to the same dimension as word embeddings. The LLM processes these interleaved embeddings in the same manner as language tokens, generating outputs in an autoregressive fashion.

During training, we freeze all the parameters of the FaceInsight model except for the adapter and graph convolutional neural network. To achieve fine-grained face perception, we adopt the binary cross entropy loss:
\begin{equation}
	\begin{aligned}
		\mathcal{L}_{bce} = & -\frac{1}{N} \sum_{i=1}^{N} \sum_{j=1}^{K} [ y_{ij} \log(P(y_{ij})) \\
		& + (1 - y_{ij}) \log(1 - P(y_{ij}))],
	\end{aligned}
\end{equation}
where $N$ is the number of training images, $K$ is the number of facial characteristics, and $P(\cdot)$ is the predicted probability for the instance.
Furthermore, we apply a logical loss $\mathcal{L}_{c}$ through the logical constraint module to ensure that logically consistent outputs are produced.
The overall objective for optimization is formulated as follows:
\begin{equation}
	\mathcal{L} = \mathcal{L}_{bce} + \mathcal{L}_{c}.
\end{equation}

\subsection{Facial Knowledge Alignment}
In the fine-grained understanding of facial images, complex correlations exist between facial characteristics, which we term facial knowledge. This knowledge can be classified into two types: weak and strong. Weak knowledge refers to uncertain relationships, where the presence of characteristic A in a face suggests that characteristic B may occur with varying probabilities. Strong knowledge, conversely, pertains to definite relationships: when characteristic A is present, characteristic B will either certainly appear or be entirely excluded. To effectively model these two types of correlations, we have developed two distinct modules, each tailored to address one of these categories.

\noindent \textbf{Correlation Constraint Module}
To effectively capture the relationships among facial characteristics, we introduce an correlation constraint module. Specifically, we first construct a initial correlation map to encode the interdependencies between facial characteristics. Next, we employ a graph convolutional network (GCN) to progressively embed these dependencies into the model's text feature space.
To model the dependency of facial characteristics in the form of conditional probabilities, i.e., $p(y_j|y_i)$, which represents the probability of  characteristic $y_j$ occurring given the presence of characteristic $y_i$, we observe that the resulting correlation map is asymmetric. To construct this correlation map, we first calculate the co-occurrence frequencies of characteristic pairs in the training set, resulting in a matrix $M \in {R}^{N \times N}$. Here, $N$ denotes the number of facial characteristics, and $m_{ij}$ represents the co-occurrence frequency of $y_i$ and $y_j$. Using this co-occurrence matrix,  the conditional probability matrix $P=(p_{ij})_{N \times N}$ can then be derived as follows:
\begin{equation}
	p_{ij} = \frac{m_{ij}}{n_i},
\end{equation}
where $n_i$ represents the occurrence frequency of $y_i$ in the training set, and $p_{ij}$ denotes the probability of $y_j$ appearing when $y_i$ appears.

Overfitting the matrix to the training set can negatively impact generalization. To mitigate this, we set a threshold to filter out noisy edges.
Specifically, for each $p_{ij}$, we retain only those elements greater than or equal to a threshold  $\tau$, setting the remaining elements to zero.   This process yields a sparse matrix $C=(c_{ij})_{N \times N}$ as follows:
\begin{equation}
	c_{ij} =
	\begin{cases}
		1, & \text{if  }  p_{ij} \geq \tau, \\
		0, & \text{if  }  p_{ij} < \tau.
	\end{cases}
\end{equation}

When applying the text features, the features of facial characteristics are computed as the weighted sum of its own features and those of other characteristics. However, a potential issue with this approach is over-smoothing, where features become so homogenized that features from other characteristics become indistinguishable. To address this, we adopt the following reweighting scheme to mitigate the over-smoothing of textual representations:
\begin{equation}
	\hat{c}_{ij} =
	\begin{cases}
		1-\omega, & \text{if } i = j, \\
		\frac{\omega}{\sum_{j=1, i\neq j}^N c_{ij}} \times c_{ij}, & \text{if } i \neq j,
	\end{cases}
	\label{weak_correlation}
\end{equation}
where $\omega$ is a hyperparameter that controls the weighting between the node's own features and those of other characteristics. By adjusting $\omega$ a fixed weight is assigned to itself during feature updates, while the weights for other characteristics are determined by their distribution within the neighborhood. As $\omega$ approaches 1, the node's own features are effectively disregarded, whereas when $\omega$ approaches 0, the influence of other characteristics is minimized.

Let the adjacency matrix be denoted as $\hat{C}=(\hat{c}_{ij})_{n \times n}$. The matrix $\hat{C}$ emphasizes the importance of the characteristic itself and weights other characteristics according to their relationships.
Specifically, guided by the correlation prior $\hat{C}$, we employ $L$ GCN layers to progressively refine the input features $T_p^0$, where $T_p^0$ is the feature ensemble of the user prompts.
The update for the $l$-th GCN layer is defined as follows:
\begin{equation}
	T_p^l = \rho(\hat{C}T_p^{l-1} W^{l-1}),
\end{equation}
where $W$ is a learnable parameter matrix and $\rho$ is a non-linear function. The final optimized textual representations are obtained through a residual connection, given by $\hat{T}_p=T_p^0 + T_p^L$. Here, $\hat{T}_p$ incorporates both the initial features and the progressively refined features from the $L$ GCN layers, facilitating accurate perception.

\noindent \textbf{Logical Constraint Module}
The correlation constraint module focus on capturing the uncertain correlations between facial characteristics, and ignores the absolutely certain correlations. Therefore, we further equip a logical constraint module to learn these certain correlations among facial characteristics.
This module introduces an additional logical loss during training, which is derived from a set of logical rules used to describe these determined dependencies. Specifically, we calculate the loss for violations of integrity constraints to force the model to make logically consistent perception.
For example, if "bald" is described as a characteristic of an individual, other hair-related characteristics, such as "bangs" or "wavy hair," should not be described. This is a logical rule, which encoded in first-order logic:
\[
\forall x \, \text{Bald}(x) \rightarrow (\neg \text{Bangs}(x) \land \neg \text{WavyHair}(x)).
\]

During the training process, 22 rules like this one are included to form mutually exclusive relationships. There are also other types of constraints, such as for certain groups, logical rules require that at least one of each group must be positive. we evaluate the probability of such constraint being violated, yielding our expected logical loss.
We formulate this loss as:
\begin{equation}
	\mathcal{L}_{c} = \frac{1}{N} \sum_{i=1}^{N}  \sum_{j=1}^{M} \mathcal{V}_i(\text{rule}_{j}) \cdot P(y_{ij}),
	\label{strong_correlation}
\end{equation}
where $N$ is the number of training images, $M$ is the number of rules, $\mathcal{V}_i(\cdot)$ computing the probability of violating the rules, and $P(\cdot)$is the probability of the preceding characteristic in each rule.

\section{Experiment}

This section presents a comprehensive evaluation of our proposed FaceInsight across three critical face perception tasks: (1) attribute recognition, (2) age, gender, and race estimation, and (3) facial expression prediction.
Considering the cost of computational resources, we conduct both training-free and fine-tuning experiments using benchmark datasets and provide comparative analyses with nine existing large multimodal models.

\subsection{Tasks and Datasets}
Each task is designed to test different dimensions of face perception, requiring models to handle semantic, demographic, and affective inference from facial imagery. We evaluate the proposed model on three face perception tasks using widely adopted datasets.

\textbf{Attribute Recognition:} We utilize the MAAD \cite{terhorst2021maad} and CelebA \cite{liu2015deep} datasets. MAAD, built on VGGFace2, comprises approximately 3.3 million facial images annotated with 47 attributes. CelebA includes over 200,000 celebrity images, each labeled with 40 binary facial attributes.

\textbf{Age/Gender/Race Estimation:} This task is evaluated on the FairFace \cite{karkkainen2021fairface} and UTKFace \cite{zhifei2017cvpr} datasets. FairFace provides balanced annotations of age, gender, and race across 108,501 images spanning seven racial groups. UTKFace consists of more than 20,000 facial images annotated with age, gender, and ethnicity, and includes substantial variation in pose, expression, illumination, occlusion, and resolution.

\textbf{Facial Expression Prediction:} We utilize the ExpW \cite{zhang2018facial} and TRAF-DB \cite{li2017reliable} datasets for expression analysis. ExpW contains 91,793 facial images annotated with one of seven basic emotions: angry, disgust, fear, happy, sad, surprise, or neutral, featuring spontaneous expressions captured in real-world settings. TRAF-DB includes approximately 15,000 images labeled with basic or compound expressions, annotated by 40 independent raters.
We adopt standard evaluation metrics for each task, including mean accuracy, recall, precision, and F1 score. The F1 score is computed as the average across individual sample-level F1 scores.

\subsection{Experimental Settings}

We compare the performance of FaceInsight against nine large multimodal models in a training-free setting. These models include InternVL2.5-8B \cite{chen2024expanding}, LLaVA-v1.5-7B \cite{liu2023visual}, LLaVA-OneVision-7B-SI \cite{li2024llava}, Mantis-SIGLIP-8B \cite{jiang2024mantis}, MiniCPM-Llama3-v2.5 \cite{yao2024minicpm}, Monkey-Chat \cite{li2024monkey}, Phi-3.5-Vision-Instruct \cite{abdin2024phi}, Qwen2.5-VL-7B-Instruct \cite{bai2025qwen2}, and NVILA-8B \cite{liu2024nvila}.
Among these, we fine-tune three representative models on task-specific datasets to enable a detailed performance comparison: LLaVA-v1.5-7B \cite{liu2023visual}, LLaVA-OneVision-7B-SI \cite{li2024llava}, and NVILA-8B \cite{liu2024nvila}.

We trained our framework simultaneously on six training datasets, using moderate batch sizes and learning rates. All remaining hyperparameters followed the configuration used in the instruction-tuning stage of LLaVA-v1.5, as described in \cite{liu2024improved}. During both training and inference, the FaceInsight model incorporated face segmentation maps generated by Pyfacer \cite{deng2020retinaface} as part of its input.
For the training-free evaluation, each model was prompted using task-specific textual templates. For instance, in the attribute recognition task, prompts followed the pattern: \textit{“Analyze the facial features in this image.”} The models’ generated outputs were parsed and mapped to the binary attribute space.
We fine-tuned selected models on each task using the corresponding training set. The image encoder was optionally frozen depending on the model, and learning rates were tuned using a grid search. Training was performed using the AdamW optimizer with early stopping based on validation accuracy to prevent overfitting.

\subsection{Comparison with State-of-the-Art Methods}
We report experimental results organized by task to better highlight the performance of the proposed FaceInsight model across different dimensions of face perception. For each task, we compare the training-free and fine-tuned performance with other large multimodal models and provide insights into FaceInsight’s capabilities.

\subsubsection{Results of Face Attribute Recognition}
\begin{figure}[ht]
	\centering
	\includegraphics[width=1\columnwidth]{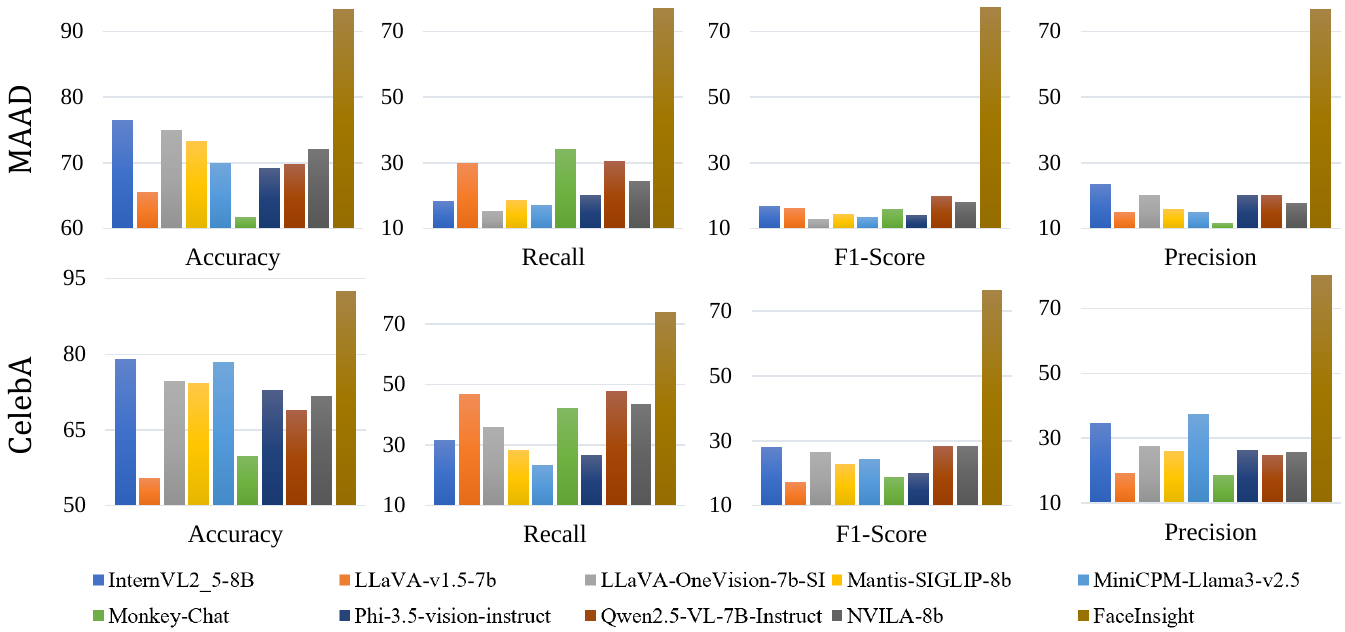}
	\caption{Performance comparison of FaceInsight and nine MLLMs on the MAAD and CelebA datasets.}
	\label{fig_results_far_mllm}
\end{figure}

\
\newline Fig. \ref{fig_results_far_mllm} presents the performance on the MAAD and CelebA datasets under the training-free setting. The experimental results show that FaceInsight outperforms existing MLLMs across all four evaluation metrics. Although individual models show reasonable accuracy, their performance in the other three metrics remains significantly low. In other words, although these models generate some accurate descriptions of facial features, they also produce a considerable amount of incorrect information, which is detrimental to practical applications. As illustrated in the visualization results on the Fig. \ref{fig_intro}, while existing MLLMs output some correct facial features, they are also prone to errors and contradictions. This highlights a critical flaw in their face perception capabilities.

Table \ref*{results_far} reports the performance of FaceInsight alongside three fine-tuned MLLMs on face attribute recognition tasks. After fine-tuning, FaceInsight continues to outperform the three fine-tuned MLLMs across all four evaluation metrics. While fine-tuning improved the performance of the MLLMs on the task-specific datasets, their results remain significantly lower than FaceInsight's across all indicators. This suggests that, despite the refinement process, these models are still not sufficiently optimized for face perception tasks. In particular, the improvement in performance through fine-tuning does not bring these models close to the required level of accuracy and reliability for practical applications in face perception.

The disparity in performance underscores the limitations of existing MLLMs, even after fine-tuning, in fully capturing the complexities of facial attributes. This performance gap underscores the need for FaceInsight, which integrate advanced face perception capabilities to address these shortcomings and deliver more reliable and nuanced results in facial attribute analysis.

\begin{table}[h]
	\caption{Performance comparison of FaceInsight with fine-tuned MLLMs on the MAAD and CelebA datasets. Bold values denote the best performance.}\vspace{-10pt}
	\resizebox{1\columnwidth}{!}{
	\begin{tabular}{cc|cccc}
		\toprule
		Datasets & Models  & Accuracy       & Recall         & F1-Score       & Precision      \\
        \midrule
		\multirow{4}{*}{MAAD}   & LLaVA-OneVision-7b-SI & 87.56          & 45.32          & 45.40          & 46.76          \\
		& LLaVA-v1.5-7b         & 83.33          & 38.13          & 36.19          & 39.96          \\
		& NVILA-8b              & 87.77          & 47.42          & 46.34          & 45.83          \\
		& \textbf{FaceInsight}           & \textbf{93.55} & \textbf{77.20} & \textbf{77.64} & \textbf{77.16} \\ \midrule
		\multirow{4}{*}{CelebA} & LLaVA-OneVision-7b-SI & 87.38          & 50.12          & 49.23          & 49.77          \\
		& LLaVA-v1.5-7b         & 81.59          & 33.23          & 30.83          & 39.90          \\
		& NVILA-8b              & 87.32          & 50.81          & 49.67          & 49.29          \\
		& \textbf{FaceInsight}           & \textbf{92.58} & \textbf{74.28} & \textbf{76.68} & \textbf{80.64} \\
        \bottomrule
	\end{tabular}
}
	\label{results_far}
    \vspace{-12pt}
\end{table}





\subsubsection{Results of Age/Gender/Race Estimation}
\begin{figure}[!b]
	\centering
	\includegraphics[width=1\columnwidth]{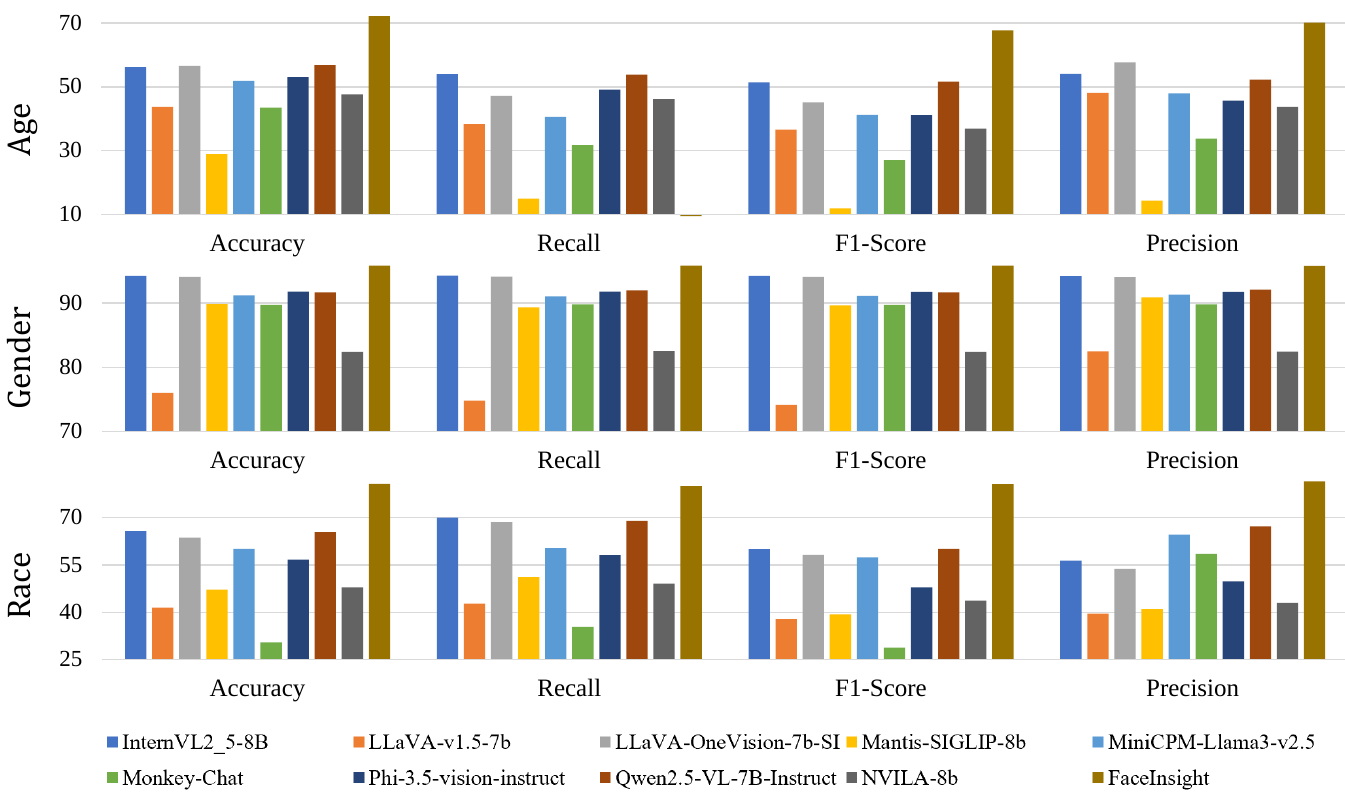}
	\caption{Performance comparison of FaceInsight and nine MLLMs on the FairFace dataset.}
	\label{fig_results_ef}
\end{figure}

\begin{figure}[!b]
	\centering
	\includegraphics[width=1\columnwidth]{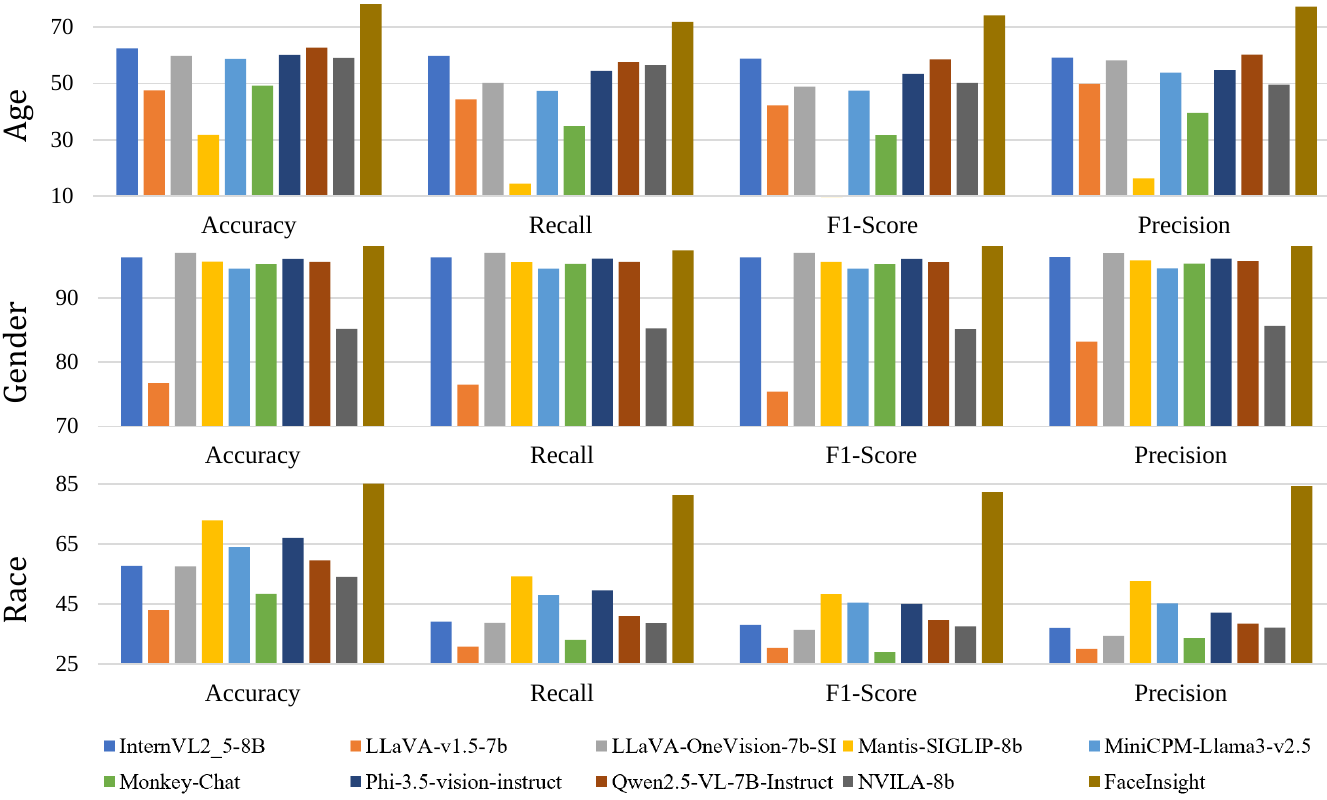}
	\caption{Performance comparison of FaceInsight and nine MLLMs on the UTKFace dataset.}
	\label{fig_results_eu}
\end{figure}

\
\newline Fig. \ref{fig_results_ef} and \ref{fig_results_eu} present the performance of FaceInsight and nine MLLMs on the FairFace and UTKFace datasets. FaceInsight consistently outperforms all other models across all evaluation metrics in the fine-grained analysis of the three demographic attributes.
The results indicate that existing MLLMs show significant improvement over their performance in previous attribute recognition tasks, particularly in gender prediction. This enhancement is likely due to the binary classification nature of the gender task, which simplifies the problem and reduces its complexity. However, there remains considerable room for improvement in race and age prediction tasks. Accurately predicting race from facial features requires models capable of capturing more nuanced visual details—an area where current MLLMs still face challenges.

Tables \ref{results_ef} and \ref{results_eu} present the performance of FaceInsight and three fine-tuned MLLMs on this task. Similar to the previous task, while fine-tuning improves the performance of the MLLMs, they still fall short compared to our proposed method. Specifically, FaceInsight performs similarly to the three comparison models in gender and race estimation, a result largely due to the relative simplicity of these tasks. However, FaceInsight significantly outperforms the fine-tuned MLLMs in the more complex task of age estimation. This demonstrates that our method more effectively leverages facial knowledge for fine-grained analysis.
While existing MLLMs show improvements in demographic attribute evaluation, FaceInsight outperforms them across all performance indicators. However, there remains significant potential to enhance these models' ability to accurately predict race and further refine their capabilities in demographic analysis.

\begin{table}[b]
	\caption{Performance comparison of FaceInsight with fine-tuned MLLMs on the FairFace dataset. Bold values denote the best performance.}\vspace{-12pt}
	\resizebox{1\columnwidth}{!}{
    \begin{tabular}{cc|cccc}
    	\toprule
    	Attributes    & Models                 & Accuracy       & Recall         & F1-Score       & Precision      \\ \midrule
    	\multirow{4}{*}{Age}    & LLaVA-OneVision-7b-SI & 61.87          & 54.98 & 56.46 & 58.81          \\
    	& LLaVA-v1.5-7b         & 52.49          & 45.12          & 47.61          & 49.73          \\
    	& NVILA-8b              & 61.60          & 53.38          & 55.71          & 59.61          \\
    	& \textbf{FaceInsight}  & \textbf{72.12} & \textbf{66.20}          & \textbf{67.67}          & \textbf{70.05} \\ \midrule
    	\multirow{4}{*}{Gender} & LLaVA-OneVision-7b-SI & 95.18          & 95.14          & 95.17          & 95.21          \\
    	& LLaVA-v1.5-7b         & 93.33          & 92.13          & 92.56          & 93.04          \\
    	& NVILA-8b              & 94.69          & 94.71          & 94.68          & 94.67          \\
    	& \textbf{FaceInsight}  & \textbf{95.79} & \textbf{95.80} & \textbf{95.78} & \textbf{95.76} \\ \midrule
    	\multirow{4}{*}{Race}   & LLaVA-OneVision-7b-SI & 76.97          & 74.43          & 76.58          & \textbf{83.47} \\
    	& LLaVA-v1.5-7b         & 51.33          & 55.21          & 49.21          & 58.78          \\
    	& NVILA-8b              & 79.99          & 79.77          & 79.77          & 79.88          \\
    	& \textbf{FaceInsight}  & \textbf{80.66} & \textbf{79.95} & \textbf{80.54} & 81.42          \\ \bottomrule
    \end{tabular}
    \vspace{-12pt}
}
\label{results_ef}
\end{table}

\begin{table}[b]
	\caption{Performance comparison of FaceInsight with fine-tuned MLLMs on the UTKFace dataset. Bold values denote the best performance.}\vspace{-12pt}
	\resizebox{1\columnwidth}{!}{
	\begin{tabular}{cc|cccc}
		\toprule
		Attributes    & Models                 & Accuracy       & Recall         & F1-Score       & Precision      \\ \midrule
		\multirow{4}{*}{Age}    & LLaVA-OneVision-7b-SI & 65.73          & 60.57 & 61.12          & 62.36          \\
		& LLaVA-v1.5-7b         & 54.51          & 49.43          & 50.17          & 51.87          \\
		& NVILA-8b              & 64.06          & 59.78          & 60.02          & 61.53          \\
		& \textbf{FaceInsight}  & \textbf{78.16} &  \textbf{71.85}         & \textbf{74.15} & \textbf{77.26} \\ \midrule
		\multirow{4}{*}{Gender} & LLaVA-OneVision-7b-SI & 97.16          & 97.16          & 97.16          & 97.17          \\
		& LLaVA-v1.5-7b         & 97.01          & 96.98          & 96.88          & 96.79          \\
		& NVILA-8b              & 97.43          & \textbf{97.43} & 97.43          & 97.43          \\
		& \textbf{FaceInsight}  & \textbf{98.43} & 97.36          & \textbf{98.07} & \textbf{98.16} \\ \midrule
		\multirow{4}{*}{Race}   & LLaVA-OneVision-7b-SI & 85.56          & \textbf{83.44} & 81.54          & 81.66          \\
		& LLaVA-v1.5-7b         & 67.44          & 43.77          & 42.36          & 44.32          \\
		& NVILA-8b              & 88.06          & 78.77          & 79.52          & 83.55          \\
		& \textbf{FaceInsight}  & \textbf{88.66} & 81.21          & \textbf{82.19} & \textbf{84.20} \\ \bottomrule
	\end{tabular}
}
\label{results_eu}
\end{table}

Additionally, Table \ref{results_e} compares the performance of the proposed FaceInsight method with two unified face perception models, FaceXFormer \cite{narayan2024facexformer}, and FairFace \cite{karkkainen2021fairface}, across three fine-grained face attribute perception tasks. The results demonstrate a substantial performance improvement with FaceInsight, underscoring its effectiveness in enhancing face attribute perception.
Compared with facial representation learning methods, FaRL and FaceXFormer are pre-trained models designed to obtain universal facial representations for various downstream tasks. These models benefit from the transfer learning capabilities gained through large-scale image-text pair pre-training and demonstrate considerable performance in face attribute perception. However, they fall short in capturing the complex correlations among face attributes, which is where our method excels. We further explore the impact of high-order correlation in FaceInsight in the ablation experiment section.

\begin{table}[!tb]
	\centering
	\caption{Performance comparison of FaceInsight with facial representation learning methods on the FairFace and UTKFace datasets. Bold values denote the best performance.}\vspace{-12pt}
	\resizebox{0.7\columnwidth}{!}{
    	\begin{tabular}{cc|cc}
    		\toprule
    		Attributes & Models       & FairFace       & UTKFace        \\  \midrule
    		\multirow{3}{*}{Age}  & FairFace    & 59.70          & 61.70          \\
    		& FaceXFormer & 59.38          & 63.93          \\
    		& \textbf{FaceInsight} & \textbf{72.12} & \textbf{78.16} \\  \midrule
    		\multirow{3}{*}{Gen}  & FairFace    & 94.20          & 92.50          \\
    		& FaceXFormer & 95.20          & 95.69          \\
    		& \textbf{FaceInsight} & \textbf{95.79} & \textbf{98.43} \\   \midrule
    		\multirow{3}{*}{Race} & FairFace    & 78.01          & 83.90          \\
    		& FaceXFormer & 77.91          & 87.45          \\
    		& \textbf{FaceInsight} & \textbf{80.66} & \textbf{88.66}  \\
    		\bottomrule
    	\end{tabular}
    }
	\label{results_e}
    \vspace{-12pt}
\end{table}

\subsubsection{Results of Facial Expression Recognition}
\begin{figure}[!h]
	\centering
	\includegraphics[width=1\columnwidth]{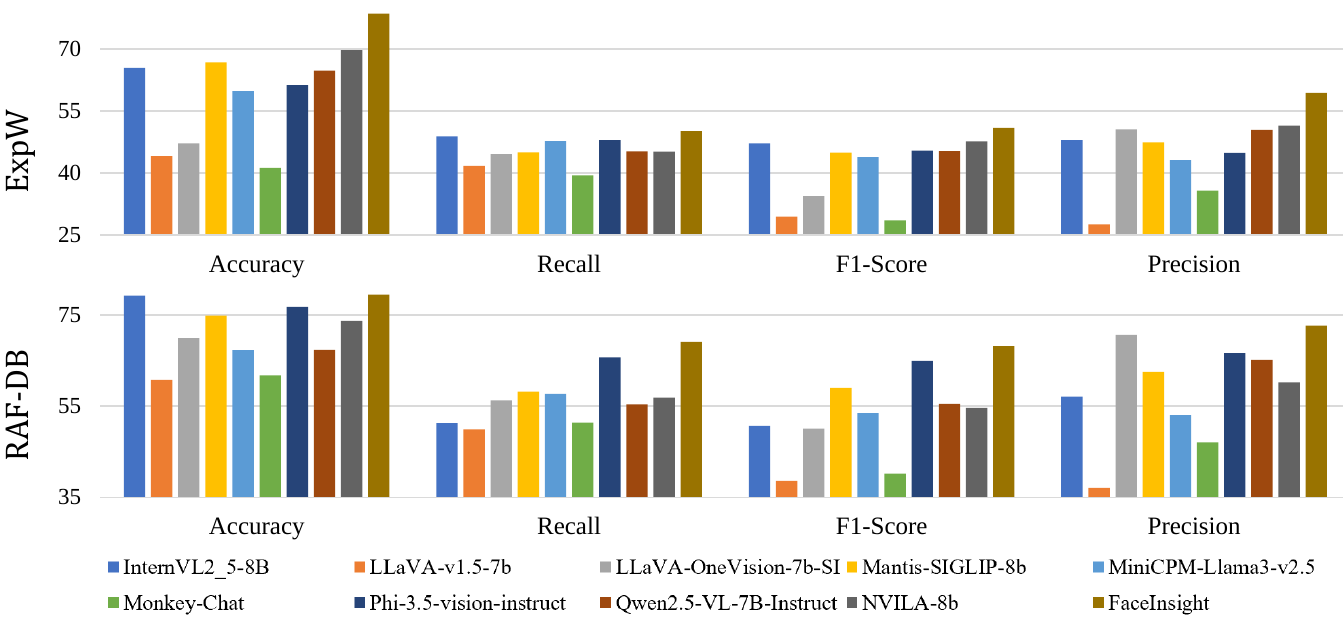}
	\caption{Performance comparison of FaceInsight and nine MLLMs on the ExpW and RAF-DB datasets.}
	\label{fig_results_exp}
\end{figure}

\
\newline Fig. \ref{fig_results_exp} and Table \ref{results_exp} present the performance of FaceInsight and existing MLLMs on two facial expression datasets under both the training-free and fine-tuned settings. Across all four evaluation metrics, FaceInsight consistently achieves the highest performance. In the training-free setting, the accuracy of existing MLLMs in facial expression recognition typically ranges between 50\% and 60\%. Although fine-tuning on expression-specific datasets improves accuracy to approximately 70\%, this remains notably lower than the performance achieved by FaceInsight. Moreover, the other three metrics—recall, precision, and F1 score—also show significant gaps when compared to our proposed method. Similar to previous face perception tasks, these models may attain moderate accuracy, but their low recall and precision values indicate a limited ability to consistently and correctly identify expressions in real-world scenarios. This highlights  the effectiveness of FaceInsight in delivering more reliable and contextually accurate expression analysis.

\begin{table}[h]
	\caption{Performance comparison of FaceInsight with fine-tuned MLLMs on the ExpW and RAF-DB datasets. Bold values denote the best performance.}\vspace{-12pt}
    \resizebox{1\columnwidth}{!}{
	\begin{tabular}{cccccc}
		\toprule
		Datasets & Models                 & Accuracy       & Recall         & F1-Score       & Precision      \\ \midrule
		\multirow{4}{*}{ExpW}   & LLaVA-OneVision-7b-SI & 71.68          & 47.59          & 49.23          & 56.33          \\
		& LLaVA-v1.5-7b         & 70.72          & 46.71          & 48.14          & 55.42          \\
		& NVILA-8b              & 74.31          & 48.36          & 49.52          & 56.68          \\
		& \textbf{FaceInsight}  & \textbf{78.47} & \textbf{50.09} & \textbf{50.92} & \textbf{59.36} \\ \midrule
		\multirow{4}{*}{RAF-DB} & LLaVA-OneVision-7b-SI & 78.43          & 60.94          & 60.34          & 66.94          \\
		& LLaVA-v1.5-7b         & 77.45          & 62.60          & 61.01          & 65.68          \\
		& NVILA-8b              & 76.40          & 57.18          & 57.22          & 63.66          \\
		& \textbf{FaceInsight}  & \textbf{79.47} & \textbf{69.11} & \textbf{68.19} & \textbf{72.68} \\ \bottomrule
	\end{tabular}
}
\label{results_exp}
\vspace{-12pt}
\end{table}


\subsection{Visualization Results}

\begin{figure}[!ht]
	\centering
	\includegraphics[width=1\columnwidth]{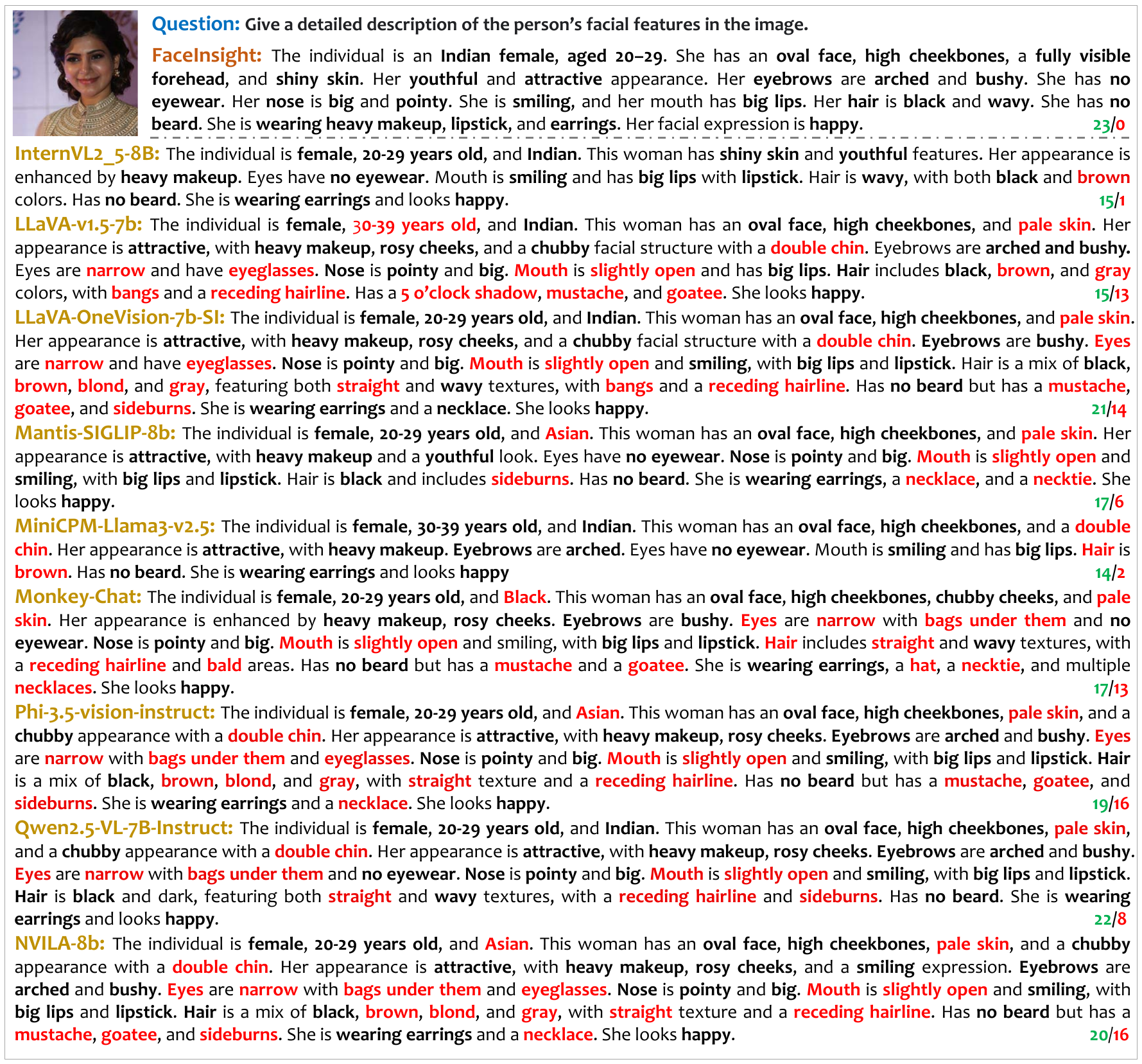}
	\caption{Qualitative results for FaceInsight and all MLLMs. Red numbers indicate incorrect facial descriptions, while green numbers indicate correct ones. }
	\label{fig_results_v1}
\end{figure}

Fig. \ref{fig_results_v1} presents qualitative visualization results comparing all MLLMs and FaceInsight in real-world face perception scenarios. In the figure, red text indicates perceptual errors, including outputs that represent logical conflicts, and hallucination phenomena. As previously discussed, applying MLLMs directly to face perception tasks often results in misinterpretations due to the dense and complex facial information compressed within a limited visual area. These limitations are clearly evident in the outputs of several MLLMs, where generated descriptions frequently diverge from the actual visual content.
We observe that existing MLLMs perform particularly poorly, often generating multiple conflicting descriptions for a single attribute—for example, simultaneously indicating both bangs and a receding hairline.
In contrast, FaceInsight exhibits significantly improved perceptual accuracy, producing coherent and visually grounded descriptions that reflect a more precise understanding of facial features in real-world conditions.

\subsection{Ablation Studies}
In the experiments, we conducted comprehensive ablation studies on different components of the proposed framework to assess their contributions. The FaceInsight framework integrates three main components: face segmentation maps, an attribute association module, and a logical constraint module. Specifically, we used LLaVA-v1.5 as a baseline with standard fine-tuning settings and employed binary cross-entropy loss for the final prediction of each attribute. This baseline is denoted as "org," with the following components added sequentially: "org+seg" indicates the face segmentation maps, "org+seg+corr" incorporates the attribute association module, and "org+seg+corr+lgc" represents the full framework with the addition of the logical constraint module.

\begin{figure}[htbp]
	\centering
	\begin{subfigure}[b]{0.23\textwidth}
		\includegraphics[width=\textwidth]{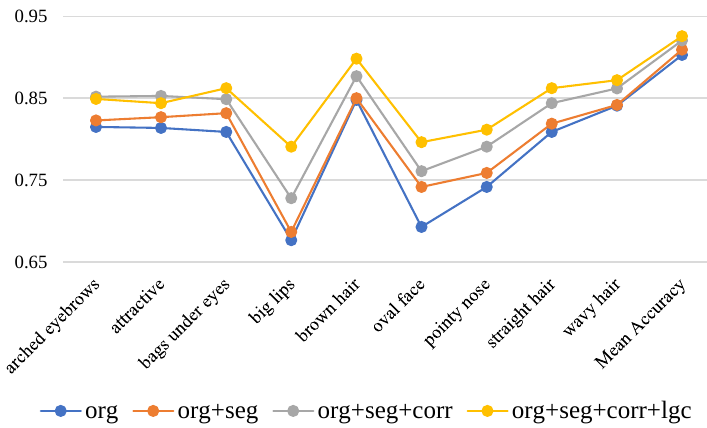}
		\caption{Impact of key components.}
		\label{fig:sub1}
	\end{subfigure}
	\begin{subfigure}[b]{0.23\textwidth}
		\includegraphics[width=\textwidth]{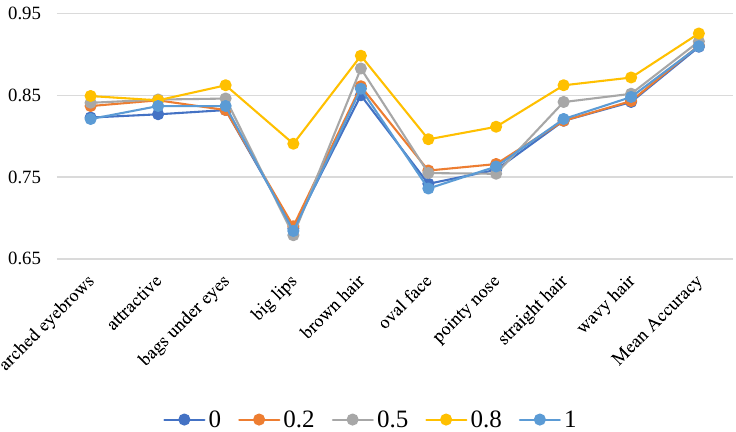}
		\caption{Impact of $\omega$ on accuracy.}
		\label{fig:sub2}
	\end{subfigure}
    \vspace{-12pt}
	\caption{Ablation study results for FaceInsight.}
\end{figure}

The ablation results are presented in Fig. \ref{fig:sub1}, highlighting several key observations. On the CelebA dataset, each proposed component consistently improves performance. The FaceInsight significantly outperforms the baseline in both single attribute accuracy and average accuracy, achieving an average mAP improvement of 2.3\%, which demonstrates the effectiveness and superiority of the proposed method. Specifically, adding the segmentation modalities (denoted as "org+seg") improves performance by 0.82\% over the baseline ("org"), particularly enhancing attributes such as "pointy nose," "oval face," and "arched eyebrows," which illustrates the benefit of integrating face segmentation maps for learning geometric features. The incorporation of high-order attribute correlations leads to an overall improvement of 1.48\%, demonstrating the advantage of integrating face attribute correlation knowledge into the multimodal LLM optimization process. Augmented by the attribute association module, our method achieves a 0.95\% improvement in mAP. This enhancement reflects the module's capability to capture potential dependencies among attributes. Additionally, the performance gains from the logical constraint module demonstrate the effectiveness of logistic loss in regularizing the definitive correlations among facial attributes.


To investigate the impact of different values of $\omega$ in Eq.\ref{weak_correlation} on attribute accuracy, we evaluated $\omega$ across the set {0,0.2,0.5,0.8,1}, as shown in Fig. \ref{fig:sub2}. The figure illustrates the significance of balancing the weights between the attribute itself and other attributes when updating the text feature in FaceInsight. Our experiments determined the optimal value of $\omega=0.8$ through cross-validation, with
yielding the best performance on CelebA. If $\omega$ is too small, the feature fails to capture sufficient information from correlated attributes, while if $\omega$ is too large, it risks losing its own distinctive features.

\section{Conclusions}
In this work, we introduce FaceInsight, a novel framework for face perception that leverages the advanced visual representation learning capabilities of multimodal LLMs. To address the technical bottlenecks that constrain current face perception, we enhance key components within the multimodal LLMs. Specifically, we develop a correlation constraint module and a logic constraint module to capture complex facial knowledge, addressing both uncertain dependencies and deterministic relationships among facial features. Additionally, we incorporate face segmentation maps to introduce localized visual information, embedding spatially-aware facial knowledge into the MLLM framework.
Our model demonstrates strong performance across various datasets and face perception tasks, showcasing its effectiveness and superiority.
This work presents a comprehensive study of face perception based on MLLMs, emphasizing the challenges posed by the complex information perception in limited visual areas and offering targeted solutions. While focused on facial analysis, our approach is broadly applicable to other fine-grained visual perception tasks.

\bibliographystyle{ACM-Reference-Format}
\bibliography{sample-base}

\begin{thebibliography}{10}

\bibitem{narayan2025facexbench}
Kartik Narayan, Vibashan VS, and Vishal~M Patel.
\newblock Facexbench: Evaluating multimodal llms on face understanding.
\newblock {\em arXiv preprint arXiv:2501.10360}, 2025.

\bibitem{sun2024face}
Haomiao Sun, Mingjie He, Tianheng Lian, Hu~Han, and Shiguang Shan.
\newblock Face-mllm: A large face perception model.
\newblock {\em arXiv preprint arXiv:2410.20717}, 2024.

\bibitem{zhao2025favchat}
Fufangchen Zhao, Ming Li, Linrui Xu, Wenhao Jiang, Jian Gao, and Danfeng Yan.
\newblock Favchat: Unlocking fine-grained facail video understanding with
  multimodal large language models.
\newblock {\em arXiv preprint arXiv:2503.09158}, 2025.

\bibitem{xing2024emo}
Bohao Xing, Zitong Yu, Xin Liu, Kaishen Yuan, Qilang Ye, Weicheng Xie, Huanjing
  Yue, Jingyu Yang, and Heikki K{\"a}lvi{\"a}inen.
\newblock Emo-llama: Enhancing facial emotion understanding with instruction
  tuning.
\newblock {\em arXiv preprint arXiv:2408.11424}, 2024.

\bibitem{awadalla2023openflamingo}
Anas Awadalla, Irena Gao, Josh Gardner, Jack Hessel, Yusuf Hanafy, Wanrong Zhu,
  Kalyani Marathe, Yonatan Bitton, Samir Gadre, Shiori Sagawa, et~al.
\newblock Openflamingo: An open-source framework for training large
  autoregressive vision-language models.
\newblock {\em arXiv preprint arXiv:2308.01390}, 2023.

\bibitem{liu2023visual}
Haotian Liu, Chunyuan Li, Qingyang Wu, and Yong~Jae Lee.
\newblock Visual instruction tuning.
\newblock {\em Advances in neural information processing systems},
  36:34892--34916, 2023.

\bibitem{liu2024llava}
Haotian Liu, Chunyuan Li, Yuheng Li, Bo~Li, Yuanhan Zhang, Sheng Shen, and
  Yong~Jae Lee.
\newblock Llava-next: Improved reasoning, ocr, and world knowledge, 2024.

\bibitem{li2025visual}
Yifan Li, Zhixin Lai, Wentao Bao, Zhen Tan, Anh Dao, Kewei Sui, Jiayi Shen,
  Dong Liu, Huan Liu, and Yu~Kong.
\newblock Visual large language models for generalized and specialized
  applications.
\newblock {\em arXiv preprint arXiv:2501.02765}, 2025.

\bibitem{radford2021learning}
Alec Radford, Jong~Wook Kim, Chris Hallacy, Aditya Ramesh, Gabriel Goh,
  Sandhini Agarwal, Girish Sastry, Amanda Askell, Pamela Mishkin, Jack Clark,
  et~al.
\newblock Learning transferable visual models from natural language
  supervision.
\newblock In {\em International conference on machine learning}, pages
  8748--8763. PMLR, 2021.

\bibitem{jain2024vcoder}
Jitesh Jain, Jianwei Yang, and Humphrey Shi.
\newblock Vcoder: Versatile vision encoders for multimodal large language
  models.
\newblock In {\em Proceedings of the IEEE/CVF Conference on Computer Vision and
  Pattern Recognition}, pages 27992--28002, 2024.

\bibitem{chiang2023vicuna}
Wei-Lin Chiang, Zhuohan Li, Zi~Lin, Ying Sheng, Zhanghao Wu, Hao Zhang, Lianmin
  Zheng, Siyuan Zhuang, Yonghao Zhuang, Joseph~E Gonzalez, et~al.
\newblock Vicuna: An open-source chatbot impressing gpt-4 with 90\%* chatgpt
  quality.
\newblock {\em See https://vicuna. lmsys. org (accessed 14 April 2023)},
  2(3):6, 2023.

\bibitem{qi2024cogcom}
Ji~Qi, Ming Ding, Weihan Wang, Yushi Bai, Qingsong Lv, Wenyi Hong, Bin Xu, Lei
  Hou, Juanzi Li, Yuxiao Dong, et~al.
\newblock Cogcom: Train large vision-language models diving into details
  through chain of manipulations.
\newblock {\em arXiv preprint arXiv:2402.04236}, 2024.

\bibitem{touvron2023llama}
Hugo Touvron, Thibaut Lavril, Gautier Izacard, Xavier Martinet, Marie-Anne
  Lachaux, Timoth{\'e}e Lacroix, Baptiste Rozi{\`e}re, Naman Goyal, Eric
  Hambro, Faisal Azhar, et~al.
\newblock Llama: Open and efficient foundation language models.
\newblock {\em arXiv preprint arXiv:2302.13971}, 2023.

\bibitem{zhu2023minigpt}
Deyao Zhu, Jun Chen, Xiaoqian Shen, Xiang Li, and Mohamed Elhoseiny.
\newblock Minigpt-4: Enhancing vision-language understanding with advanced
  large language models.
\newblock {\em arXiv preprint arXiv:2304.10592}, 2023.

\bibitem{li2023blip}
Junnan Li, Dongxu Li, Silvio Savarese, and Steven Hoi.
\newblock Blip-2: Bootstrapping language-image pre-training with frozen image
  encoders and large language models.
\newblock In {\em International conference on machine learning}, pages
  19730--19742. PMLR, 2023.

\bibitem{instructblip}
Wenliang Dai, Junnan Li, Dongxu Li, Anthony Meng~Huat Tiong, Junqi Zhao,
  Weisheng Wang, Boyang Li, Pascale Fung, and Steven Hoi.
\newblock Instructblip: Towards general-purpose vision-language models with
  instruction tuning, 2023.

\bibitem{ye2023mplug}
Qinghao Ye, Haiyang Xu, Guohai Xu, Jiabo Ye, Ming Yan, Yiyang Zhou, Junyang
  Wang, Anwen Hu, Pengcheng Shi, Yaya Shi, et~al.
\newblock mplug-owl: Modularization empowers large language models with
  multimodality.
\newblock {\em arXiv preprint arXiv:2304.14178}, 2023.

\bibitem{liu2024improved}
Haotian Liu, Chunyuan Li, Yuheng Li, and Yong~Jae Lee.
\newblock Improved baselines with visual instruction tuning.
\newblock In {\em Proceedings of the IEEE/CVF Conference on Computer Vision and
  Pattern Recognition}, pages 26296--26306, 2024.

\bibitem{chen2024lion}
Gongwei Chen, Leyang Shen, Rui Shao, Xiang Deng, and Liqiang Nie.
\newblock Lion: Empowering multimodal large language model with dual-level
  visual knowledge.
\newblock In {\em Proceedings of the IEEE/CVF Conference on Computer Vision and
  Pattern Recognition}, pages 26540--26550, 2024.

\bibitem{chen2023minigpt}
Jun Chen, Deyao Zhu, Xiaoqian Shen, Xiang Li, Zechun Liu, Pengchuan Zhang,
  Raghuraman Krishnamoorthi, Vikas Chandra, Yunyang Xiong, and Mohamed
  Elhoseiny.
\newblock Minigpt-v2: large language model as a unified interface for
  vision-language multi-task learning.
\newblock {\em arXiv preprint arXiv:2310.09478}, 2023.

\bibitem{lin2024vila}
Ji~Lin, Hongxu Yin, Wei Ping, Pavlo Molchanov, Mohammad Shoeybi, and Song Han.
\newblock Vila: On pre-training for visual language models.
\newblock In {\em Proceedings of the IEEE/CVF Conference on Computer Vision and
  Pattern Recognition}, pages 26689--26699, 2024.

\bibitem{lin2024moe}
Bin Lin, Zhenyu Tang, Yang Ye, Jiaxi Cui, Bin Zhu, Peng Jin, Jinfa Huang, Junwu
  Zhang, Yatian Pang, Munan Ning, et~al.
\newblock Moe-llava: Mixture of experts for large vision-language models.
\newblock {\em arXiv preprint arXiv:2401.15947}, 2024.

\bibitem{chen2024expanding}
Zhe Chen, Weiyun Wang, Yue Cao, Yangzhou Liu, Zhangwei Gao, Erfei Cui, Jinguo
  Zhu, Shenglong Ye, Hao Tian, Zhaoyang Liu, et~al.
\newblock Expanding performance boundaries of open-source multimodal models
  with model, data, and test-time scaling.
\newblock {\em arXiv preprint arXiv:2412.05271}, 2024.

\bibitem{yao2024minicpm}
Yuan Yao, Tianyu Yu, Ao~Zhang, Chongyi Wang, Junbo Cui, Hongji Zhu, Tianchi
  Cai, Haoyu Li, Weilin Zhao, Zhihui He, et~al.
\newblock Minicpm-v: A gpt-4v level mllm on your phone.
\newblock {\em arXiv preprint arXiv:2408.01800}, 2024.

\bibitem{jiang2024mantis}
Dongfu Jiang, Xuan He, Huaye Zeng, Cong Wei, Max Ku, Qian Liu, and Wenhu Chen.
\newblock Mantis: Interleaved multi-image instruction tuning.
\newblock {\em arXiv preprint arXiv:2405.01483}, 2024.

\bibitem{li2024monkey}
Zhang Li, Biao Yang, Qiang Liu, Zhiyin Ma, Shuo Zhang, Jingxu Yang, Yabo Sun,
  Yuliang Liu, and Xiang Bai.
\newblock Monkey: Image resolution and text label are important things for
  large multi-modal models.
\newblock In {\em proceedings of the IEEE/CVF conference on computer vision and
  pattern recognition}, pages 26763--26773, 2024.

\bibitem{abdin2024phi}
Marah Abdin, Jyoti Aneja, Hany Awadalla, Ahmed Awadallah, Ammar~Ahmad Awan,
  Nguyen Bach, Amit Bahree, Arash Bakhtiari, Jianmin Bao, Harkirat Behl, et~al.
\newblock Phi-3 technical report: A highly capable language model locally on
  your phone.
\newblock {\em arXiv preprint arXiv:2404.14219}, 2024.

\bibitem{bai2025qwen2}
Shuai Bai, Keqin Chen, Xuejing Liu, Jialin Wang, Wenbin Ge, Sibo Song, Kai
  Dang, Peng Wang, Shijie Wang, Jun Tang, et~al.
\newblock Qwen2. 5-vl technical report.
\newblock {\em arXiv preprint arXiv:2502.13923}, 2025.

\bibitem{hassanat2022deepveil}
Ahmad~BA Hassanat, Abeer~Ahmad Albustanji, Ahmad~S Tarawneh, Malek Alrashidi,
  Hani Alharbi, Mohammed Alanazi, Mansoor Alghamdi, Ibrahim~S Alkhazi, and
  VB~Surya Prasath.
\newblock Deepveil: deep learning for identification of face, gender,
  expression recognition under veiled conditions.
\newblock {\em International Journal of Biometrics}, 14(3-4):453--480, 2022.

\bibitem{yan2023spl}
Yan Yan, Ying Shu, Si~Chen, Jing-Hao Xue, Chunhua Shen, and Hanzi Wang.
\newblock Spl-net: Spatial-semantic patch learning network for facial attribute
  recognition with limited labeled data.
\newblock {\em International Journal of Computer Vision}, 131(8):2097--2121,
  2023.

\bibitem{wu2023logical}
Haiyu Wu, Grace Bezold, Aman Bhatta, and Kevin~W Bowyer.
\newblock Logical consistency and greater descriptive power for facial hair
  attribute learning.
\newblock In {\em Proceedings of the IEEE/CVF Conference on Computer Vision and
  Pattern Recognition}, pages 8588--8597, 2023.

\bibitem{wu2023logicnet}
Haiyu Wu, Sicong Tian, Huayu Li, and Kevin~W Bowyer.
\newblock Logicnet: A logical consistency embedded face attribute learning
  network.
\newblock {\em arXiv preprint arXiv:2311.11208}, 2023.

\bibitem{canal2022survey}
Felipe~Zago Canal, Tobias~Rossi M{\"u}ller, Jhennifer~Cristine Matias,
  Gustavo~Gino Scotton, Antonio~Reis de~Sa~Junior, Eliane Pozzebon, and
  Antonio~Carlos Sobieranski.
\newblock A survey on facial emotion recognition techniques: A state-of-the-art
  literature review.
\newblock {\em Information Sciences}, 582:593--617, 2022.

\bibitem{cheng2023semi}
Zebang Cheng, Yuxiang Lin, Zhaoru Chen, Xiang Li, Shuyi Mao, Fan Zhang, Daijun
  Ding, Bowen Zhang, and Xiaojiang Peng.
\newblock Semi-supervised multimodal emotion recognition with expression mae.
\newblock In {\em Proceedings of the 31st ACM International Conference on
  Multimedia}, pages 9436--9440, 2023.

\bibitem{li2024facial}
Yifan Li, Anh Dao, Wentao Bao, Zhen Tan, Tianlong Chen, Huan Liu, and Yu~Kong.
\newblock Facial affective behavior analysis with instruction tuning.
\newblock In {\em European Conference on Computer Vision}, pages 165--186,
  2024.

\bibitem{cao2020rank}
Wenzhi Cao, Vahid Mirjalili, and Sebastian Raschka.
\newblock Rank consistent ordinal regression for neural networks with
  application to age estimation.
\newblock {\em Pattern Recognition Letters}, 140:325--331, 2020.

\bibitem{li2021learning}
Wanhua Li, Xiaoke Huang, Jiwen Lu, Jianjiang Feng, and Jie Zhou.
\newblock Learning probabilistic ordinal embeddings for uncertainty-aware
  regression.
\newblock In {\em Proceedings of the IEEE/CVF conference on computer vision and
  pattern recognition}, pages 13896--13905, 2021.

\bibitem{kuprashevich2023mivolo}
Maksim Kuprashevich and Irina Tolstykh.
\newblock Mivolo: Multi-input transformer for age and gender estimation.
\newblock In {\em International Conference on Analysis of Images, Social
  Networks and Texts}, pages 212--226. Springer, 2023.

\bibitem{ranjan2017hyperface}
Rajeev Ranjan, Vishal~M Patel, and Rama Chellappa.
\newblock Hyperface: A deep multi-task learning framework for face detection,
  landmark localization, pose estimation, and gender recognition.
\newblock {\em IEEE transactions on pattern analysis and machine intelligence},
  41(1):121--135, 2017.

\bibitem{ranjan2017all}
Rajeev Ranjan, Swami Sankaranarayanan, Carlos~D Castillo, and Rama Chellappa.
\newblock An all-in-one convolutional neural network for face analysis.
\newblock In {\em 2017 12th IEEE international conference on automatic face \&
  gesture recognition (FG 2017)}, pages 17--24. IEEE, 2017.

\bibitem{qin2023swinface}
Lixiong Qin, Mei Wang, Chao Deng, Ke~Wang, Xi~Chen, Jiani Hu, and Weihong Deng.
\newblock Swinface: a multi-task transformer for face recognition, expression
  recognition, age estimation and attribute estimation.
\newblock {\em IEEE Transactions on Circuits and Systems for Video Technology},
  2023.

\bibitem{narayan2024facexformer}
Kartik Narayan, Vibashan VS, Rama Chellappa, and Vishal~M Patel.
\newblock Facexformer: A unified transformer for facial analysis.
\newblock {\em arXiv preprint arXiv:2403.12960}, 2024.

\bibitem{sun2024task}
Haomiao Sun, Mingjie He, Shiguang Shan, Hu~Han, and Xilin Chen.
\newblock Task-adaptive q-face.
\newblock {\em arXiv preprint arXiv:2405.09059}, 2024.

\bibitem{qin2025faceptor}
Lixiong Qin, Mei Wang, Xuannan Liu, Yuhang Zhang, Wei Deng, Xiaoshuai Song,
  Weiran Xu, and Weihong Deng.
\newblock Faceptor: A generalist model for face perception.
\newblock In {\em European Conference on Computer Vision}, pages 240--260.
  Springer, 2025.

\bibitem{zheng2022general}
Yinglin Zheng, Hao Yang, Ting Zhang, Jianmin Bao, Dongdong Chen, Yangyu Huang,
  Lu~Yuan, Dong Chen, Ming Zeng, and Fang Wen.
\newblock General facial representation learning in a visual-linguistic manner.
\newblock In {\em Proceedings of the IEEE/CVF conference on computer vision and
  pattern recognition}, pages 18697--18709, 2022.

\bibitem{bulat2022pre}
Adrian Bulat, Shiyang Cheng, Jing Yang, Andrew Garbett, Enrique Sanchez, and
  Georgios Tzimiropoulos.
\newblock Pre-training strategies and datasets for facial representation
  learning.
\newblock In {\em European Conference on Computer Vision}, pages 107--125.
  Springer, 2022.

\bibitem{li2022label2label}
Wanhua Li, Zhexuan Cao, Jianjiang Feng, Jie Zhou, and Jiwen Lu.
\newblock Label2label: A language modeling framework for multi-attribute
  learning.
\newblock In {\em European Conference on Computer Vision}, pages 562--579.
  Springer, 2022.

\bibitem{zhao2023prompting}
Zengqun Zhao and Ioannis Patras.
\newblock Prompting visual-language models for dynamic facial expression
  recognition.
\newblock {\em arXiv preprint arXiv:2308.13382}, 2023.

\bibitem{foteinopoulou2024emoclip}
Niki~Maria Foteinopoulou and Ioannis Patras.
\newblock Emoclip: A vision-language method for zero-shot video facial
  expression recognition.
\newblock In {\em 2024 IEEE 18th International Conference on Automatic Face and
  Gesture Recognition (FG)}, pages 1--10. IEEE, 2024.

\bibitem{deng2020retinaface}
Jiankang Deng, Jia Guo, Evangelos Ververas, Irene Kotsia, and Stefanos
  Zafeiriou.
\newblock Retinaface: Single-shot multi-level face localisation in the wild.
\newblock In {\em Proceedings of the IEEE/CVF Conference on Computer Vision and
  Pattern Recognition}, pages 5203--5212, 2020.

\bibitem{terhorst2021maad}
Philipp Terh{\"o}rst, Daniel F{\"a}hrmann, Jan~Niklas Kolf, Naser Damer,
  Florian Kirchbuchner, and Arjan Kuijper.
\newblock Maad-face: A massively annotated attribute dataset for face images.
\newblock {\em IEEE Transactions on Information Forensics and Security},
  16:3942--3957, 2021.

\bibitem{liu2015deep}
Ziwei Liu, Ping Luo, Xiaogang Wang, and Xiaoou Tang.
\newblock Deep learning face attributes in the wild.
\newblock In {\em Proceedings of the IEEE international conference on computer
  vision}, pages 3730--3738, 2015.

\bibitem{karkkainen2021fairface}
Kimmo Karkkainen and Jungseock Joo.
\newblock Fairface: Face attribute dataset for balanced race, gender, and age
  for bias measurement and mitigation.
\newblock In {\em Proceedings of the IEEE/CVF winter conference on applications
  of computer vision}, pages 1548--1558, 2021.

\bibitem{zhifei2017cvpr}
Song~Yang Zhang, Zhifei and Hairong Qi.
\newblock Age progression/regression by conditional adversarial autoencoder.
\newblock In {\em IEEE Conference on Computer Vision and Pattern Recognition
  (CVPR)}. IEEE, 2017.

\bibitem{zhang2018facial}
Zhanpeng Zhang, Ping Luo, Chen~Change Loy, and Xiaoou Tang.
\newblock From facial expression recognition to interpersonal relation
  prediction.
\newblock {\em International Journal of Computer Vision}, 126:550--569, 2018.

\bibitem{li2017reliable}
Shan Li, Weihong Deng, and JunPing Du.
\newblock Reliable crowdsourcing and deep locality-preserving learning for
  expression recognition in the wild.
\newblock In {\em Proceedings of the IEEE conference on computer vision and
  pattern recognition}, pages 2852--2861, 2017.

\bibitem{li2024llava}
Bo~Li, Yuanhan Zhang, Dong Guo, Renrui Zhang, Feng Li, Hao Zhang, Kaichen
  Zhang, Peiyuan Zhang, Yanwei Li, Ziwei Liu, et~al.
\newblock Llava-onevision: Easy visual task transfer.
\newblock {\em arXiv preprint arXiv:2408.03326}, 2024.

\bibitem{liu2024nvila}
Zhijian Liu, Ligeng Zhu, Baifeng Shi, Zhuoyang Zhang, Yuming Lou, Shang Yang,
  Haocheng Xi, Shiyi Cao, Yuxian Gu, Dacheng Li, et~al.
\newblock Nvila: Efficient frontier visual language models.
\newblock {\em arXiv preprint arXiv:2412.04468}, 2024.

\end{thebibliography}



\end{document}